\documentclass[journal]{IEEEtran}

\usepackage{lineno}
\usepackage{graphicx}
\usepackage{subfigure}
\usepackage{amssymb}
\usepackage{epstopdf}
\usepackage{booktabs}
\usepackage{caption}
\usepackage{wrapfig}
\usepackage{makecell}
\usepackage{CJK}
\usepackage{amsmath}
\usepackage{txfonts}
\usepackage{bm}
\usepackage{color}
\usepackage{subfigure}
\usepackage{multicol}
\usepackage{multirow}
\usepackage[font=small,labelsep=period ]{caption}

\begin{document}

\title{Soft-ranking Label Encoding for Robust Facial Age Estimation}

\author{Xusheng~Zeng,
Changxing~Ding,
Yonggang~Wen,
and~Dacheng~Tao

\thanks{X. Zeng and C. Ding are with the School of Electronic
and Information Engineering, South China University of Technology, 381
Wushan Road, Tianhe District, Guangzhou 510000, China (e-mail:
201721011423@mail.scut.edu.cn; chxding@scut.edu.cn).}
\thanks{Y. Wen is with the School of Computer Science and
Engineering, Nanyang Technological University, Singapore 639798 (e-mail:
ygwen@ntu.edu.sg).}
\thanks{D. Tao is with the UBTECH Sydney Artificial Intelligence Centre and the School of Computer Science,
in the Faculty of Engineering, at the University of Sydney, 6 Cleveland St., Darlington, NSW 2008,
Australia (email: dacheng.tao@sydney.edu.au).}
}

\maketitle

\begin{abstract}
Automatic facial age estimation can be used in a wide range of real-world applications. However, this process is challenging due to the randomness and slowness of the aging process.
Accordingly, in this paper, we propose a comprehensive framework aimed at overcoming the challenges associated with facial age estimation.
First, we propose a novel age encoding method, referred to as ‘Soft-ranking’, which encodes two important properties of facial age,
$\emph{i.e.}$, the ordinal property and the correlation between adjacent ages. Therefore, Soft-ranking provides a richer supervision signal for training deep models.
Moreover, we also carefully analyze existing evaluation protocols for age estimation,
finding that the overlap in identity between the training and testing sets affects the relative performance of different age encoding methods.
Finally, since existing face databases for age estimation are generally small, deep models tend to suffer from an overfitting problem. To address this issue, we propose a novel regularization strategy to encourage deep models to learn more robust features from facial parts for age estimation purposes.
Extensive experiments indicate that the proposed techniques improve the age estimation performance;
moreover, we achieve state-of-the-art performance on the three most popular age databases, $\emph{i.e.}$, Morph II, CLAP2015, and CLAP2016.
\end{abstract}


\section{Introduction}

\IEEEPARstart{A}{ge} is one of the most important facial attributes. Automatic facial age estimation can be used in a wide range of real-world applications;
for example, human computer interaction \cite{author01}, precise advertising \cite{author02}, age-based face retrieval \cite{author05}, and video surveillance \cite{author03}.
The two most common facial age estimation tasks are real age estimation and apparent age estimation.
In this paper, we aim to handle both of these tasks using a unified framework.

Although significant efforts have been devoted to age estimation \cite{author77, author78,author06,author17,author38}, it remains a challenging problem for two main reasons.
First, the aging of human faces is a random and complicated process affected by a number of internal and external factors, such as genes and living conditions.
This means that facial appearance may vary dramatically among different subjects of the same age.
Moreover, the aging process is slow, meaning that differences in facial appearance between adjacent ages of the same subject tend to be imperceptible.
Second, existing face databases for age estimation are generally small, as face images with accurate age labels are difficult to collect;
therefore, a severe overfitting problem tends to arise among existing age estimation algorithms.

\begin{figure}[tbp]
\centering
\includegraphics[width=3.5in]{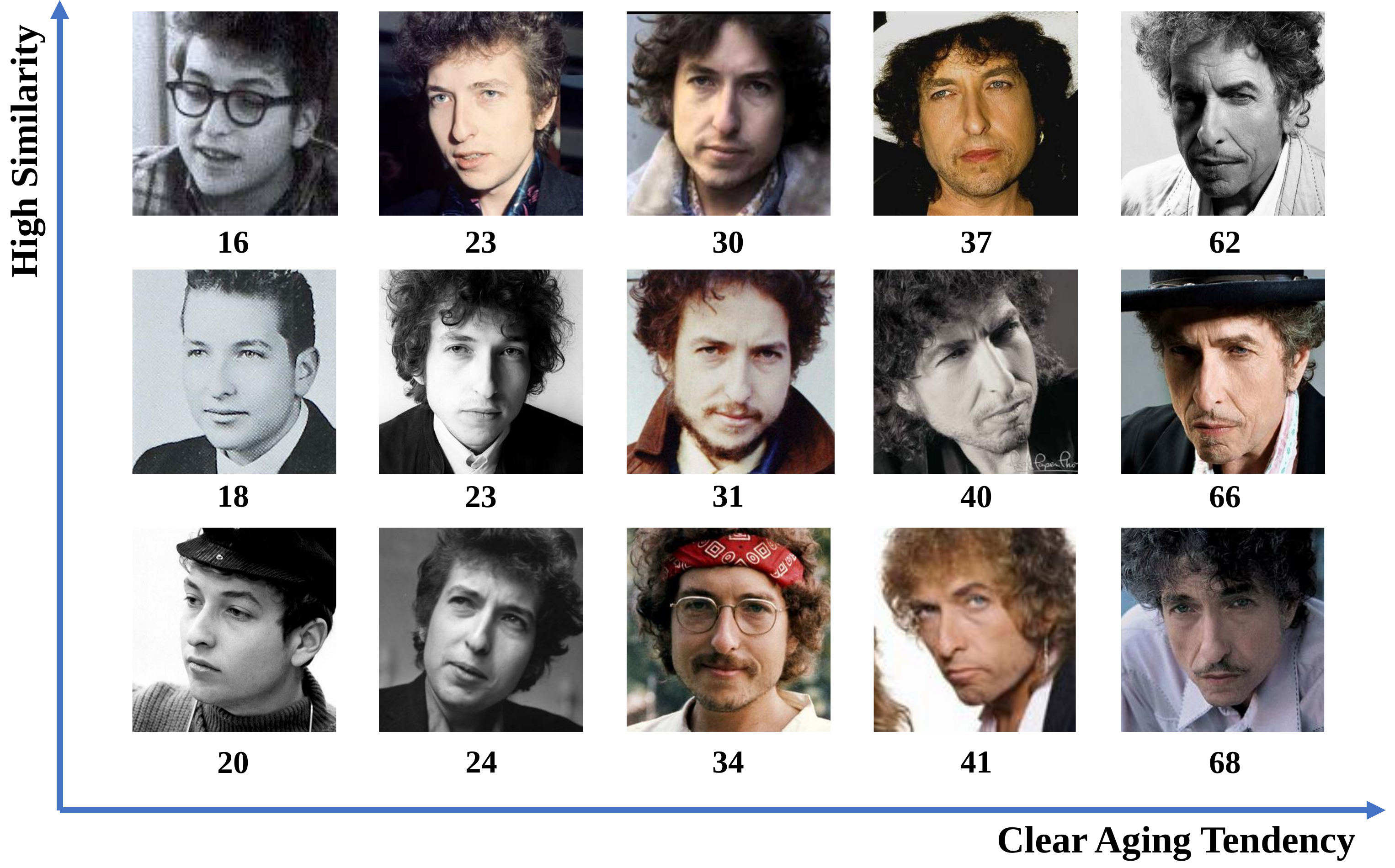}
\caption{Faces of different age values. All the above images are from the same subject.
The number below each image denotes the ground-truth age of the face.
When comparing two faces with a large age difference between them, it is easy to tell which face is older; by contrast, faces with small age gaps between them are very similar in appearance. Therefore, it is reasonable to consider both the ordinal property and correlation among adjacent ages for age encoding.
}
\label{introduction}
\end{figure}

A number of works that promote the performance of facial age estimation have been presented.
For example, many existing works have proposed new age encoding methods, which are utilized as the supervision signal for the training of deep models.
Age encoding is typically based on one property of facial age, $\emph{e.g.}$, the ordinal property \cite{author12,author13} or the correlation between adjacent ages \cite{author04, author11,author36}. However, there are no existing works that combine both properties into one unified age encoding strategy. Moreover, other works \cite{author17, author08,author40,author76} have attempted to provide complementary information to the holistic facial image, $\emph{e.g.}$, image patches, that can be utilized to improve age estimation accuracy; this information is fused at the score- \cite{author76} or feature-level \cite{author17, author08,author40}. The downside of this strategy is that it results in additional computational cost at both the training and testing stages.

Accordingly, in this paper, we approach the challenges in facial age estimation via combining multiple complementary pieces of information into one unified framework.
More specifically, we propose a novel age encoding method, which we have named `Soft-ranking'.
Each element in the encoded vector indicates the probability of a particular face being younger than a specific age;
in other words, Soft-ranking naturally takes ordinal information into consideration.
Furthermore, unlike existing ranking-based methods, the elements in Soft-ranking are soft labels; therefore, the proposed method also encodes the correlation among adjacent ages.
To the best of our knowledge, this is the first age encoding method to encode both of these facial age properties.

Moreover, inspired by the observation that it is significantly more challenging to predict facial age using an image patch than it is to do so using the holistic image,
we propose an easy-to-implement regularization method (named `Maskout') to relieve the overfitting problem of deep models.
Maskout adds multiple auxiliary tasks that employ partial feature maps of the main task for age estimation.
Therefore, we expect that the model can estimate facial age not only from the holistic image but also from an image patch.
Since these auxiliary tasks are more difficult, they play the role of regularization to the main task.
Furthermore, Maskout is only employed in the training stage and is removed during testing, giving Maskout superior computational efficiency compared with existing methods.

The main contributions of this work can be summarized as follows:

\begin{itemize}
\item We propose a novel age encoding method, named Soft-ranking, that simultaneously encodes both the ordinal information and the correlation between adjacent ages;
\item We introduce a new patch-based regularization method to reduce the risk of overfitting for deep models, which is easy to implement and incurs no extra cost at the testing stage;
\item We carefully analyze existing age estimation evaluation protocols, thereby proving empirically that the overlap in identity between the training and testing sets gives rise to misleading results during the evaluation of different age encoding methods;
\item Finally, the proposed framework achieves state-of-the-art results on three popular datasets for facial age estimation: Morph II \cite{author58}, CLAP2015 \cite{author47}, and CLAP2016 \cite{author51}. We also report promising result on the new AgeDB database \cite{author91}.
\end{itemize}

The remainder of this paper is organized as follows. Sec. \ref{sectionRelated Works} briefly reviews representative works on facial age estimation and other related topics. The proposed framework for age estimation is detailed in Sec. \ref{Method}. Experimental results on four datasets are reported and analyzed in Sec. \ref{sectionExperiments}. Finally, conclusions are drawn in Sec. \ref{sectionConclusion}.

\begin{figure*}
\centering
\includegraphics[width=7in]{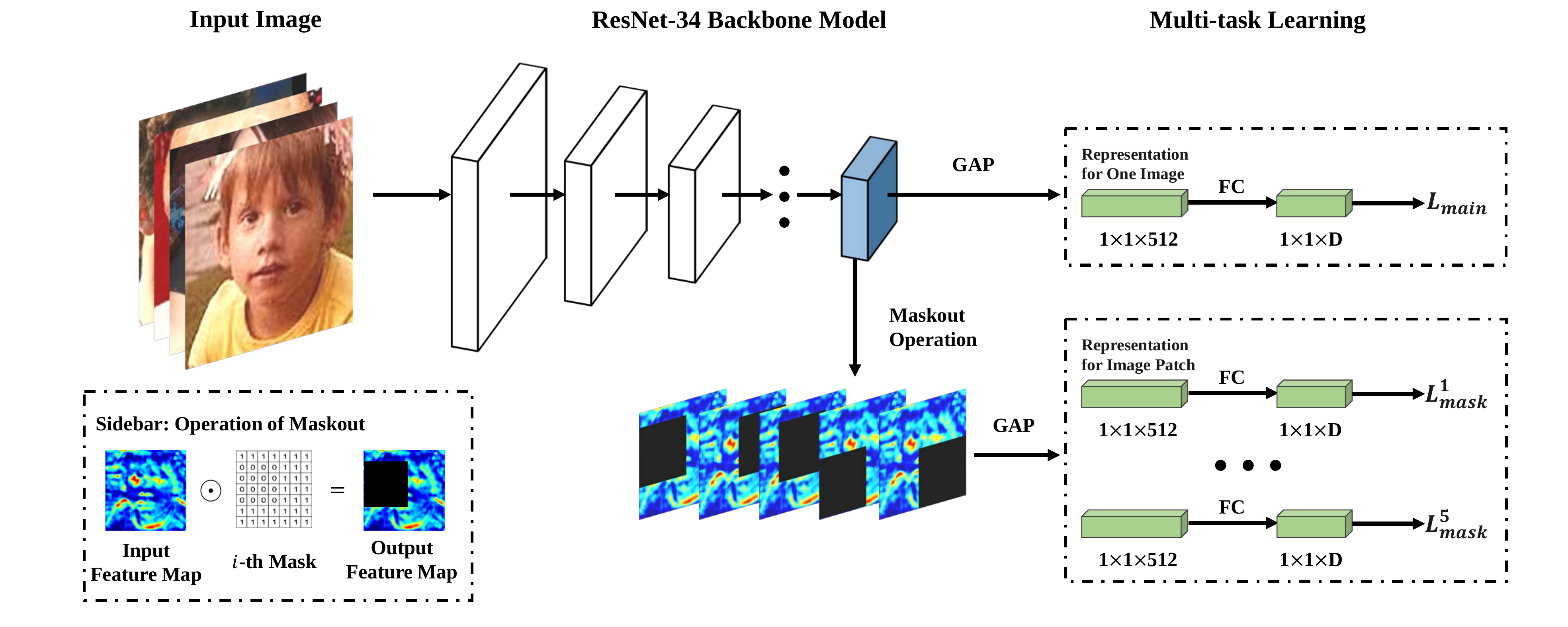}
\caption{Overview of our proposed framework for facial age estimation in the training stage.
The output feature maps generated by the final convolutional layers of ResNet-34 are fed into two types of branches,
one main branch, for facial age estimation and another five auxiliary branches, which adopt Maskout regularization.
Soft-ranking is consistently employed for all branches in this figure. At the testing stage, only the main branch is used for prediction. }
\label{pipeline}
\end{figure*}

\section{Related Works}
\label{sectionRelated Works}
In this section, we review the relevant literature in two key fields: 1) facial age estimation, and 2) multi-task learning.

\subsection{Facial Age Estimation }
Traditional methods \cite{author05, author17,author26,author67,author68,author85,author86} estimate facial age in two separate steps:
namely, feature extraction and age regression or classification.
By contrast, deep learning-based approaches \cite{author27, author28} can integrate these two individual stages into an end-to-end model that predicts the facial age directly from the raw image pixels. Due to the advantages of representation learning, deep learning-based approaches have dominated the field of age estimation in recent years.
Therefore, we will primarily review deep learning-based approaches in the following subsection.

The loss functions of early deep learning approaches tended to be based on classification \cite{author07, author88} or regression \cite{author08,author09,author87}.
However, their formulation only incorporates the ground-truth age value and ignores the relationships between all possible age values.
To overcome this problem, age encoding-based methods have been proposed, which are usually based on the property of facial age.
Among existing works, label distribution learning (LDL) \cite{author38, author36,author37,author84} and ranking \cite{author12,author13,author94} are the two most popular age encoding approaches; these methods make use of the correlation among adjacent ages and the ordinal information, respectively.
Implicitly, these two age encoding methods can provide the age classifier with more training samples \cite{author37}.
Kullback-Leibler (K-L) divergence and cross-entropy loss are usually adopted to measure the discrepancy between ground-truth and predicted age encoding vectors.
Moreover, Tan $\emph{et al.}$ \cite{author44} proposed the AGEn encoding method, which groups adjacent ages into the same category,
in order to transform the age estimation problem into a set of binary classification problems. Since adjacent ages are regarded as one category, their correlation is encoded.

Other works have adopted multiple loss functions to train deep models.
For example, Gao $\emph{et al.}$ \cite{author04} argued that a single loss function may be insufficient to train an accurate age estimation model.
Accordingly, these authors introduced an expectation loss to assist LDL: this loss is aimed at eliminating the inconsistency between the training and prediction stages of LDL-based age estimation.
For their part, Pan $\emph{et al.}$ \cite{author31} proposed a novel mean-variance loss.
Variance loss aims to minimize the variance of the predicted label distribution;
therefore, the curve of the obtained label distribution will be sharp, and the discriminative power of the model can thereby be enhanced.

\subsection{Multi-task Learning }
Multi-task learning (MTL) is a popular machine learning technique that learns multiple relevant tasks simultaneously in one model. By exploiting the correlation among tasks, MTL implicitly increases the size of training data and improves the model’s generalization ability. As a consequence, MTL has been successfully applied to many computer vision tasks, including semantic segmentation \cite{author66}, pose estimation \cite{author64}, fine-grained recognition~\cite{author81, author82}, and facial attribute estimation \cite{author63,author65,author95}.

Among the existing works exploring facial attribute estimation, Ranjan $\emph{et al.}$ \cite{author09} proposed an all-in-one network that predicts multiple facial attributes simultaneously,
including facial age. As each task in this network can benefit from the other tasks, this approach can achieve better overall performance.
Moreover, Antipov $\emph{et al.}$ \cite{author28} proposed a network for joint gender and age estimation,
an approach that can assist in improving the robustness of age estimation when the network is trained from scratch.
Several other works have designed new loss functions as auxiliary tasks for age estimation;
for example, both \cite{author04} and \cite{author31} proposed new loss functions to regularize the predicted label distributions of the age.
However, there are few works that have constructed an MTL-based model architecture dedicated to age estimation.
Accordingly, in this work, we introduce a novel MTL-based model architecture for age estimation.
Compared with existing MTL approaches in relevant topics~\cite{author09,author63,author81, author82},
all auxiliary tasks in our model are removed in the testing stage; therefore our method is very efficient during testing.

\section{Methodology}
\label{Method}

In the following, we will first briefly review two popular age encoding methods, $\emph{i.e.}$, LDL and ranking.
Our proposed Soft-ranking age encoding method will then be described in detail.
Finally, we introduce the Maskout regularization, which is designed to reduce  the overfitting problem commonly observed in deep models.

\subsection{Brief Review of LDL and Ranking}
\begin{figure*}[htbp]
\centering
\includegraphics[width=14cm]{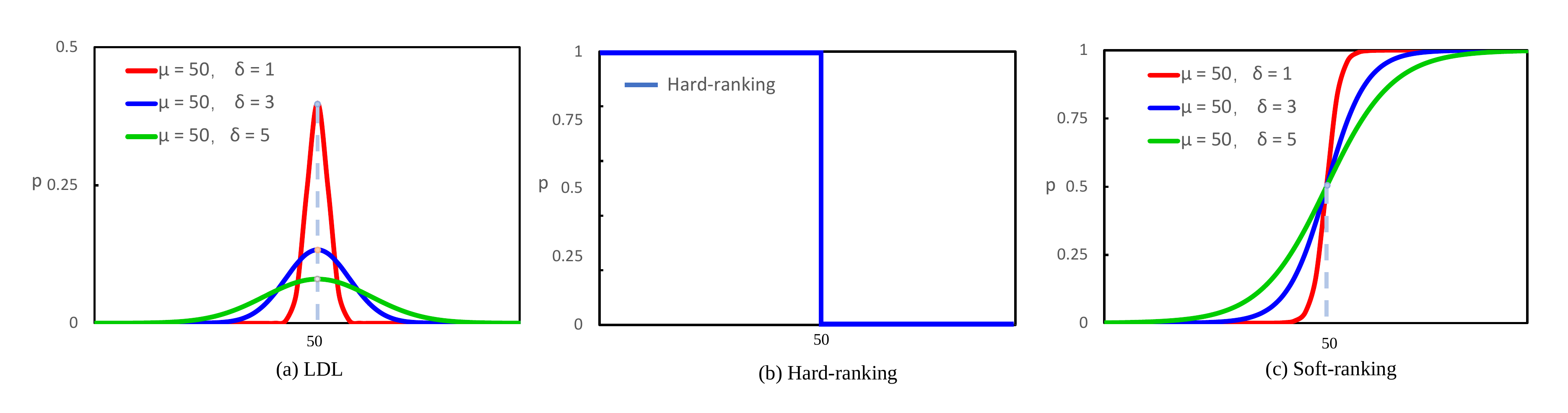}
\caption{Comparisons between different age encoding methods. The age of 50 is taken as an example. (a) LDL encodes the correlation between the ground-truth age and its adjacent ages. (b) Hard-ranking takes only the ordinal information into consideration. (c) Soft-ranking encodes both facial age properties.}
\label{three_encoding_methods}
\end{figure*}

LDL and ranking represent facial age using a vector that encodes the relation between the ground-truth age and all possible age values.
At the prediction stage, the network also predicts age encoding vectors, which are then decoded to specific age values.
Different age encoding methods are illustrated and compared in Fig. \ref{three_encoding_methods}.

\subsubsection{LDL}
Elements in a vector encoded by LDL can be regarded as the probabilities that a specific face belongs to different age value categories.
A Gaussian distribution is usually adopted to generate the ground-truth age encoding $\bm{p}_n$ for the $n$-th sample at age $y_n$.
Each element in $\bm{p}_n$ is computed as follows:

\begin{equation}
p^{k}_n = \frac{1}{\sqrt{2\pi}\sigma}\exp\left(-\frac{\left(k-y_n\right)^{2}}{2\sigma^{2}}\right),
\label{labeldistribution}
\end{equation}
where $k \in \left\{1,2,...,K\right\}$, $p^k_n$ denotes the $k$-th element of ${\bm{p}_n}$, $\sigma$ is a hyper-parameter that controls the degree of correlation between adjacent ages, and $K$ refers to the biggest age considered by the model. The age encoding vector ${\bm{p}_n}$ is then L1-normalized to ensure that the sum of its elements is equal to 1.

Moreover, LDL can be applied to Convolutional Neural Networks (CNNs) \cite{author45, author56}.
In the following, we take ResNet-34~\cite{author48} illustrated in Fig. \ref{pipeline} as an explanatory example.
The output of the Global Average Pooling (GAP) layer in ResNet-34 is utilized as the facial representations $\bm{f}\in \bm{\mathcal{R}}^{ 512}$ for an image $\bm{\mathcal{X}}$.
One fully-connected (FC) layer is attached to $\bm{f}$ in order to obtain the output of the network,
which can be denoted as $\bm{o} = {\bm{W}}^T{\bm{f}} +{\bm{b}}$; here,
$\left\{\bm{W}, \bm{b} \right\}$ are the parameters of the FC layer.
$\bm{o}\in \bm{\mathcal{R}}^{D}$ and $D$ equals to $K$ for LDL.

During the training stage, the K-L divergence is adopted to measure the discrepancy between the ground-truth and predicted age encodings:
\begin{equation}
\begin{split}
L_{ldl}=-\frac{1}{N}\sum_{n=1}^{N}\sum_{k=1}^{K} p_{n}^{k}\ln{\frac{p_{n}^{k}}{\hat{p}_{n}^{k}}}, \label{kldiv}
\end{split}
\end{equation}
where
\begin{equation}
\hat{p}_{n}^{k} = \frac{\exp{(o_n^k)}}{\sum_{j=1}^{K} \exp{(o_n^j)}}
\label{softmax}.
\end{equation}
Here, N stands for the batch size, while $\hat{p}_{n}^{k}$ and $p_n^k$ denote the $k$-th element of the predicted and the ground-truth age encoding vectors for $n$-th sample, respectively.
In the testing stage, the age value $\hat{y}_n$ of one face is obtained by calculating the expectation of the predicted age encoding vector.

\subsubsection{Hard-ranking}
Ranking-based age encoding methods represent the age of a face using an ordered rank \cite{author12}.
Each element in the encoded vector $\bm{v}_n$ refers to whether or not the face in question is older than a specific age value.
Accordingly, given the ground-truth age $y_n$, the $k$-th element in $\bm{v}_n$ is defined as:
\begin{equation}
v^k_n = \begin{cases}1 & if(y_n>k)\\0 & otherwise\end{cases},
\end{equation}
where $k \in \left\{1,2,...,K-1 \right\}$. Since $v^k_n$ is equal to either 1 or 0,
we refer to existing ranking-based approaches as  ‘Hard-ranking' to differentiate them from the proposed Soft-ranking method.
Hard-ranking methods are usually transformed into $K-1$ binary classification problems \cite{author12};
therefore, the dimension $D$ of the FC layer in Fig. \ref{pipeline} for Hard-ranking is equal to $2\times{\left(K-1\right)}$.
We apply a softmax function to every two neurons of the FC layer in order to obtain the output for each binary classifier.
Here, cross-entropy loss is adopted as the loss function for Hard-ranking.
During the testing stage, the age of one face is obtained by aggregating the predictions from all $K-1$ classifiers:
\begin{equation}
\hat{y}_n = 1+\sum_{k=1}^{K-1}\hat{v}^k_n,
\end{equation}
where $\hat{v}^k_n \in \left\{ 0, 1\right\} $ stands for the prediction made by the $k$-th binary classifier.

\subsection{Age Encoding by Soft-ranking}
\begin{figure}[htb]
\centering
\includegraphics[width=8cm]{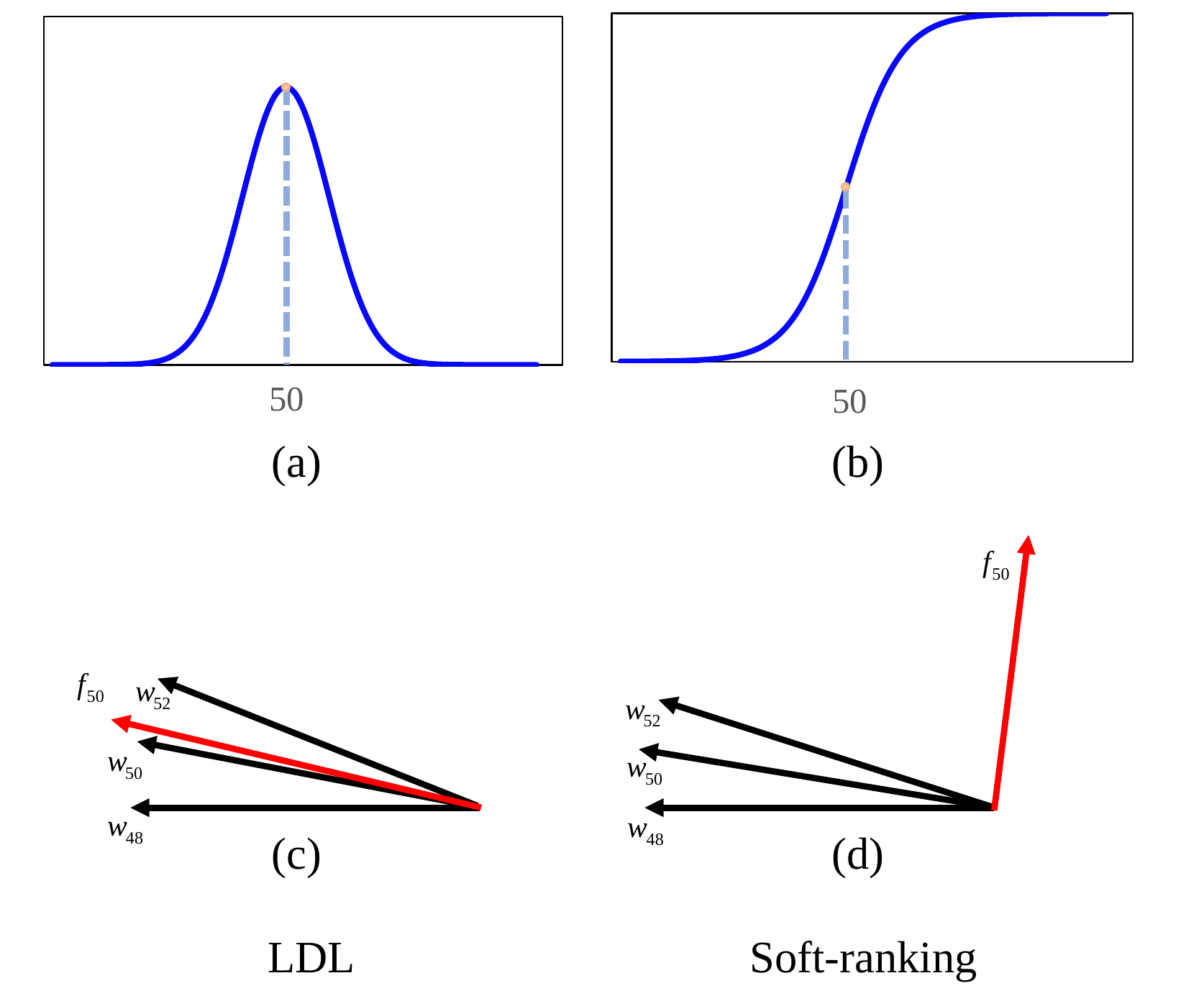}
\caption{Comparisons between two age encoding strategies: LDL and Soft-ranking.
We take the age values of 48, 50, and 52 as examples. (a) Age encoding for age 50 by LDL. (b) Age encoding for age 50 by Soft-ranking.
(c) Illustration of facial representations and classifiers for networks equipped with LDL.
(d) Illustration of facial representations and classifiers for networks equipped with Soft-ranking.
$\bm{f}_{50}$ stands for the facial representation of one face at age 50, while $\bm{w}_{48}$, $\bm{w}_{50}$, and $\bm{w}_{52}$ refer to classifiers for ages 48, 50, and 52, respectively.
All of these are simplified to vectors in the 2D space. The bias $\bm{b}$ for age classifiers is omitted in the interests of simplicity.
It is clear that the position of classifiers $\bm{w}_{48}$ and $\bm{w}_{52}$ are interchangeable from the perspective of $\bm{f}_{50}$ in LDL; by contrast,
these classifiers cannot replace each other in Soft-ranking.}
\label{introduction_1}
\end{figure}

We illustrate the limitation of LDL in Fig. \ref{introduction_1}, in which both facial representation and classifiers are simplified to vectors in 2D space. It is clear that age encodings produced by LDL are bilaterally symmetric. As outlined in Fig. \ref{introduction_1}(c), this property creates an ambiguity problem between different age classifiers $\bm{w}$, $\emph{i.e.}$, column vectors in $\bm{W}$, for networks equipped with LDL.

Accordingly, we propose the Soft-ranking age encoding strategy to address the above problem. Soft-ranking can be regarded as a combination of LDL and Hard-ranking.
Each element in a vector encoded using the Soft-ranking strategy represents the probability that a given face is younger than a specific age.
We further adopt the cumulative distribution function of the Gaussian distribution to generate the ground-truth age encoding vector ${\bm{p}_n}$ for the $n$-th image at age $y_n$.
The $k$-th element of ${\bm{p}_n}$ can be represented as:
\begin{equation}
p^{k}_n = \frac{1}{2}\left[1+erf(\frac{k-y_n}{\sqrt{2}\sigma})\right],
\label{cdf1}
\end{equation}
where

\begin{equation}
erf(x) = \frac{2}{\sqrt{\pi}}\int_{0}^{x}e^{-t^2}dt,
\end{equation}
and $k \in \left\{1,2,...,K \right\}$. $\sigma$ is a hyper-parameter that controls the degree of correlation between adjacent ages in Soft-ranking,
in a manner identical to that outlined in Eq. \ref{labeldistribution}.
It is apparent that when $k$ is equal to $y_n$, $p^{k}_n$ is equal to 0.5; when $k$ is bigger than $y_n$, $ p^{k}_n$ will be bigger than 0.5, and vice versa.

Soft-ranking offers a more comprehensive way to describe the relationship between a specific face and all possible age values.
First, Soft-ranking’s $\sigma$ parameter enables it to control the degree of correlation between adjacent ages;
second, the monotonically increasing property enables it to describe the ordinal information between all age values.
As illustrated in Fig. \ref{introduction_1}(d), the relationship between facial representations and all age classifiers is clearly defined and without ambiguity.
By contrast, an ambiguity problem exists for adjacent age classifiers in networks equipped with LDL, as illustrated in Fig. \ref{introduction_1}(c).

In the interests of fair comparison, we adopt the same network structure for the Soft- and Hard-ranking methods.
In particular, we also apply a softmax function to every two neurons of the FC layer in Fig. \ref{pipeline} in order to obtain the outputs for each classifier.
There are only two trivial and intuitive differences:
first, the number of classifiers in Soft-ranking is $K$; second, the value of $D$ is equal to $2K$.
The outputs of the $k$-th classifier for the $n$-th sample are denoted as $\hat{p}_{n}^{k0}$ and $\hat{p}_{n}^{k1}$, respectively.
Moreover, to facilitate fair comparison with LDL, we also adopt K-L divergence to measure the discrepancy between the ground-truth and predicted age encodings, as follows:
\begin{equation}
L_{sr}=-\frac{1}{2N}\sum_{n=1}^{N}\sum_{k=1}^{K} \sum_{j=0}^{1} p_{n}^{kj}\ln{\frac{p_{n}^{kj}}{\hat{p}_{n}^{kj}}},
\end{equation}
where $p_{n}^{k0}=p^{k}_n$ and $p_{n}^{k1}=1-p^{k}_n$.

It is worth noting here that when $\sigma$ is 0, $p^{k}_n$ becomes a binary label expect when $k$ is equal to $y_n$.
In this case, $L_{sr}$ is equivalent to the cross-entropy loss.
Therefore, Hard-ranking can be regarded as a special case of Soft-ranking when $\sigma$ is 0.
During the testing stage, the way that age prediction results are obtained via Soft-ranking is different from the method employed by LDL and Hard-ranking.
Formally, the age value $\hat{y}_n$ is predicted as follows:
\begin{equation}
\hat {y}_n = \arg\mathop {\min}\limits_{k}abs(\hat{p}_{n}^{k0}-\hat{p}_{n}^{k1}),
\label{estimate_cdf}
\end{equation}
where $abs(x)$ returns the absolute value of $x$. When $abs(\hat{p}_{n}^{k0}-\hat{p}_{n}^{k1})$ is at its minimum, $\hat{p}_{n}^{k0}$ achieves the closest value to 0.5. Moreover, since we use the returned value of $k$ in Eq. \ref{estimate_cdf} as the predicted age, the age values predicted by Soft-ranking are integers.

\subsection{Maskout Regularization}
\begin{figure}[tbp]
\centering
\includegraphics[width=3.5in]{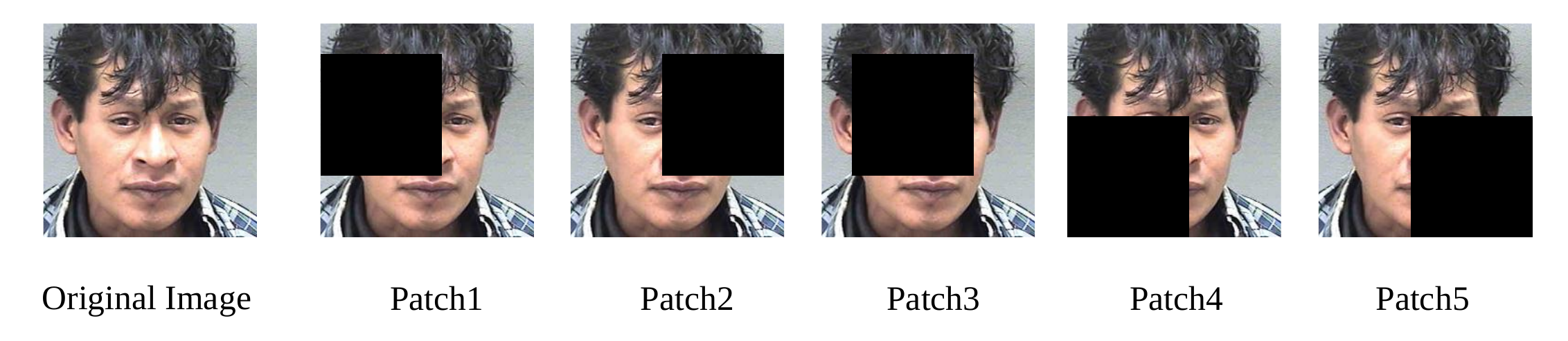}
\caption{
Strategy employed for the construction of image patches. To illustrate this more effectively, we use the original image to represent the feature maps that are output by the final convolutional layer of the backbone model.
}
\label{patch_shape}
\end{figure}

As it is difficult to label the precise age of one face, existing databases for facial age estimation are usually small.
Therefore, deep models trained on these databases tend to suffer from an overfitting problem.
One solution to relieving this overfitting problem is to find effective regularizations.
As noted above and illustrated in Fig. \ref{patch_shape}, it is significantly more challenging to predict the age value from an image patch than it is to do so from a holistic face image, as an image patch contains less information. Inspired by this observation, we propose the Maskout method to generate more difficult age estimation tasks.
In this way, deep models can be regularized, enabling them to learn more robust features for age estimation.

As illustrated in Fig. \ref{pipeline}, we add five auxiliary branches based on Maskout to the network during training.
The main task in Fig. \ref{pipeline} makes predictions based on the complete feature maps produced by the backbone model;
for their part, the five auxiliary branches make use of different types of masked feature maps for prediction.
These masked feature maps are obtained by erasing responses at pre-defined regions in the complete feature maps.

As illustrated in the sidebar of Fig. \ref{pipeline}, each mask is a matrix with the same height and width as the feature map.
Elements in the $i$-th mask $\bm{M}_i$ can be defined as follows:

\begin{equation}
M_i(x,y)= \begin{cases}0 & {
\begin{array}{c}{{ C_{x_i} -r\leq x\leq C_{x_i} +r-1}
}\\ { {C_{y_i} -r\leq y\leq C_{y_i} +r-1}}\end{array}
},

\\1 & \quad else \end{cases}
\end{equation}
where $(C_{xi}, C_{yi})$ stands for the rounded coordinates of the $i$-th predefined facial landmark in the feature map, while $r$ denotes the radius of the erased area in the mask.
We generate five masks according to the location of the five most salient facial landmarks; namely, the centers of the eyes, the tip of the nose, and the two corners of the mouth.
The generated masks can be viewed in Fig. \ref{patch_shape}.
The input feature maps for the $i$-th auxiliary branch can be obtained as follows:
\begin{equation}
\widetilde{\bm{F}}_i = \bm{F}\odot \bm{M}_i,
\end{equation}
where $\bm{F}$ stands for the feature maps output by the backbone model, while  $\odot$ represents the element-wise multiplication between the mask and each channel in $\bm{F}$.
Subsequently, a GAP layer is attached to $\widetilde{\bm{F}}_i$ to produce the branch-specific facial representation and Soft-ranking is used as the training target of the branch.

In practice, we use a set of universal masks for all face images during training.
This is because the size of the final feature maps is very small ($7\times7$ in our implementation), meaning that the location change of the five facial landmarks is negligible.

\subsection{Age Estimation Framework}
By combining the new components for age estimation introduced above, the overall loss function for the proposed age estimation framework can be represented as follows:
\begin{equation}
L = L_{main}+ \lambda \sum_{i=1}^5 L_{mask}^i,
\label{overall}
\end{equation}
where $L_{main}$ represents the loss of the main task for age estimation, while $L_{mask}^i$ denotes the loss of the $i$-th auxiliary branch that adopts Maskout regularization.
All of the above loss functions are based on Soft-ranking and realized using K-L divergence. Finally, $\lambda$ refers to the weight of the Maskout loss term.

In the testing stage, only the main task is maintained for age prediction. Therefore, all auxiliary branches play their regularization role in the training stage only. Moreover, we average the prediction results of the original image and its horizontally flipped version in testing.

\section{Experiments}
\label{sectionExperiments}
\begin{figure*}[h]
\centering
\includegraphics[width=7.2in]{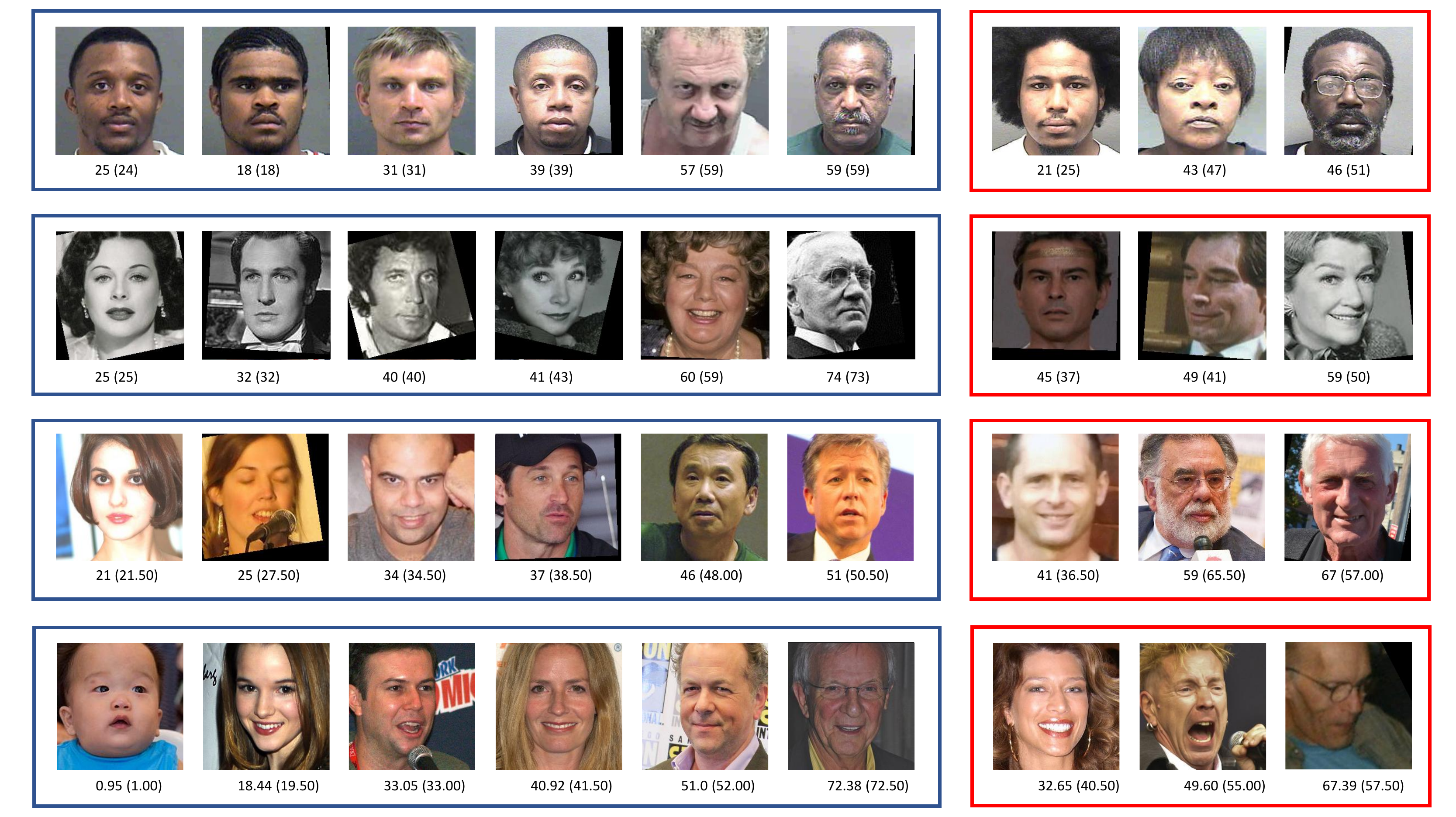}
\caption{
Examples of age estimation results obtained by the proposed approach on four datasets.
Images in the four rows are selected from Morph II, AgeDB, CLAP2015, and CLAP2016, respectively.
Sample images with high and low age prediction accuracy are framed in blue and red rectangles, respectively.
The numbers below each image refer to the ground-truth and predicted age, respectively.}
\label{samples}
\end{figure*}

In this section, we now systematically evaluate the proposed methods and compare them with the age estimation performance of state-of-the-art approaches.
Experiments are conducted on four databases: Morph II \cite{author58}, AgeDB \cite{author91}, CLAP2015 \cite{author47}, and CLAP2016 \cite{author51}.
Sample images from the four databases are presented in Fig. \ref{samples}.

\subsection{Datasets}
\textbf{Morph II} \cite{author58} is one of the largest publicly available real age databases.
It contains 55,134 images of 13,617 subjects. All images are mugshots; therefore, their image quality is generally good.
Two popular evaluation protocols are adopted: the first randomly splits (RS) \cite{author04,author11,author31} 80 percent of the images into a training set and allocates the remaining 20 percent of images to a testing set, while the second is a five-fold subject-exclusive (SE) \cite{author31} protocol. Following \cite{author31}, we use RS and SE respectively to denote the two protocols.
As there is no identity overlap in the SE protocol, it is more difficult than the RS protocol.

\textbf{AgeDB} \cite{author91} is one of the largest in-the-wild databases for real age estimation.
It contains 16,488 images of 568 subjects, with the average number of images per subject being 29.
The minimum and maximum age values are 1 and 101, respectively,  while the average age range for each subject is 50.3 years.
Moreover, identity information is available for each image; however, there is no official evaluation protocol for age estimation in this database.
In this paper, we adopt five-fold cross validation to evaluate the performance of the proposed methods.
It is worth noting that there is no identity overlap between the five folds of data.

\textbf{Chalearn LAP 2015} (CLAP2015) \cite{author47} is an apparent age estimation database that was collected for the CLAP2015 challenge.
The database contains 4,699 images in total, each of which was labeled by multiple annotators.
The label of each image is the mean age value of the annotations rather than the real age of the face.
In addition to estimated mean age value, the standard variance of the annotations for each image is also provided.
The images are split into three subsets: a training subset comprising 2,476 images, a validation subset of 1,136 images, and a testing subset containing 1,087 images. In line with existing works \cite{author06, author76, author27, author37, author44},
we merge the training and validation subsets into a larger training set and report results on the testing subset.

\textbf{Chalearn LAP 2016} (CLAP2016) \cite{author51} is a dataset that was released in 2016.
It includes 4,113 images for training, 1,500 images for validation, and 1,979 images for testing.
The method of obtaining image labels and the evaluation protocol are the same as for CLAP2015.

\subsection{Evaluation Metrics}
We adopt the most popular evaluation metric, $\emph{i.e.}$, Mean Absolute Error (MAE), to conduct the evaluation on the real age estimation task. MAE is computed as follows:
\begin{equation}
\frac{1}{N}\sum_{n=1}^N|\hat{y}_n -y_n|,
\end{equation}
where $\hat{y}_n$ and $y_n$ stand for the estimated and ground-truth age of the $n$-th image in the testing set, respectively, while $N$ is the number of images in the testing set.

For apparent age estimation, we adopt $\epsilon$-error \cite{author47} as the evaluation metric. $\epsilon$-error is computed as follows:
\begin{equation}
\frac{1}{N}\sum_{n=1}^N(1-\exp{(-\frac{(\hat{y}_n -y_n)^2}{2\sigma_n^2})}),
\end{equation}
where $\sigma_n$ denotes the standard variance of the annotations for the $n$-th testing image.
Unlike MAE, which considers the estimation error for each image equally, this metric assigns lower weights to the images with larger standard variance.

\subsection{Implementation Details}
\textbf{Data preparation.} For each face image, we use the publicly available MTCNN \cite{author49} model to detect the five most salient facial landmarks
(both eye centers, the tip of the nose, and both corners of the mouth).
An affine transformation is then estimated from the detected landmarks to align each face into an upright pose with an image size of $230 \times 230$ pixels.
Sample images after normalization are presented in Fig. \ref{samples}. In the training stage,
we employ five data augmentation strategies to relieve the overfitting problem commonly observed in deep models.
All training images are first randomly cropped to $224 \times 224$ pixels and then randomly either flipped or not flipped in the horizontal direction.
Finally, one of the following three strategies is randomly chosen and applied: random resizing and cropping, random rotation, or color jitter.

\textbf{Training Details.}
The popular ResNet-34 architecture \cite{author48} is utilized as the backbone of the model.
Since the above age estimation databases are relatively small, we pre-train the deep models in two steps:
first, we pre-train the model using the large-scale MS-Celeb-1M \cite{author50} face recognition database;
second, the obtained face recognition model can be further trained on the large-scale IMDB-WIKI database \cite{author27} for age estimation.
Finally, the pre-trained model is fine-tuned on each of the above four databases for age estimation, respectively.
In the following, pre-training with the first step only and both steps are referred to one-step and two-step pre-training strategies, respectively.

All experiments are conducted using four NVIDIA TITAN X GPUs based on the PyTorch 4.0 framework.
The stochastic gradient descent (SGD) strategy~\cite{sutskever2013importance} is used to optimize the network, with momentum set as 0.9 and weight decay set as 0.0002.
The batch size is set to 400, with each GPU processing 100 images. The total number of epochs is 100 for all databases.
The initial learning rate is set as 0.001 and decreased by a factor of 10 at the 80-th and the 90-th epoch.
In the implementation of Maskout regularization,
the auxiliary branches are added at the 10-th epoch and their parameters initialized using those of the main branch.
Moreover, we empirically set $\lambda$ in Eq. \ref{overall} as 0.3 and 0.1 for controlled (Morph II) and uncontrolled (AgeDB and CLAP) face databases, respectively;
in other words, we impose stronger regularization on controlled face databases, since their images contain less variation and are therefore more vulnerable to overfitting.

\subsection{Ablation Study}
In this section, we report on the ablation study conducted to justify the effectiveness of each component of the proposed approach.
Experiments are conducted on Morph II and AgeDB databases, where the identity information of each image is available.
The one-step pre-training strategy is adopted for clean comparison.

\subsubsection{Comparisons between different age encoding strategies (subject-exclusive protocol)}
\begin{figure}[htb]
\centering
\includegraphics[width=8.9cm]{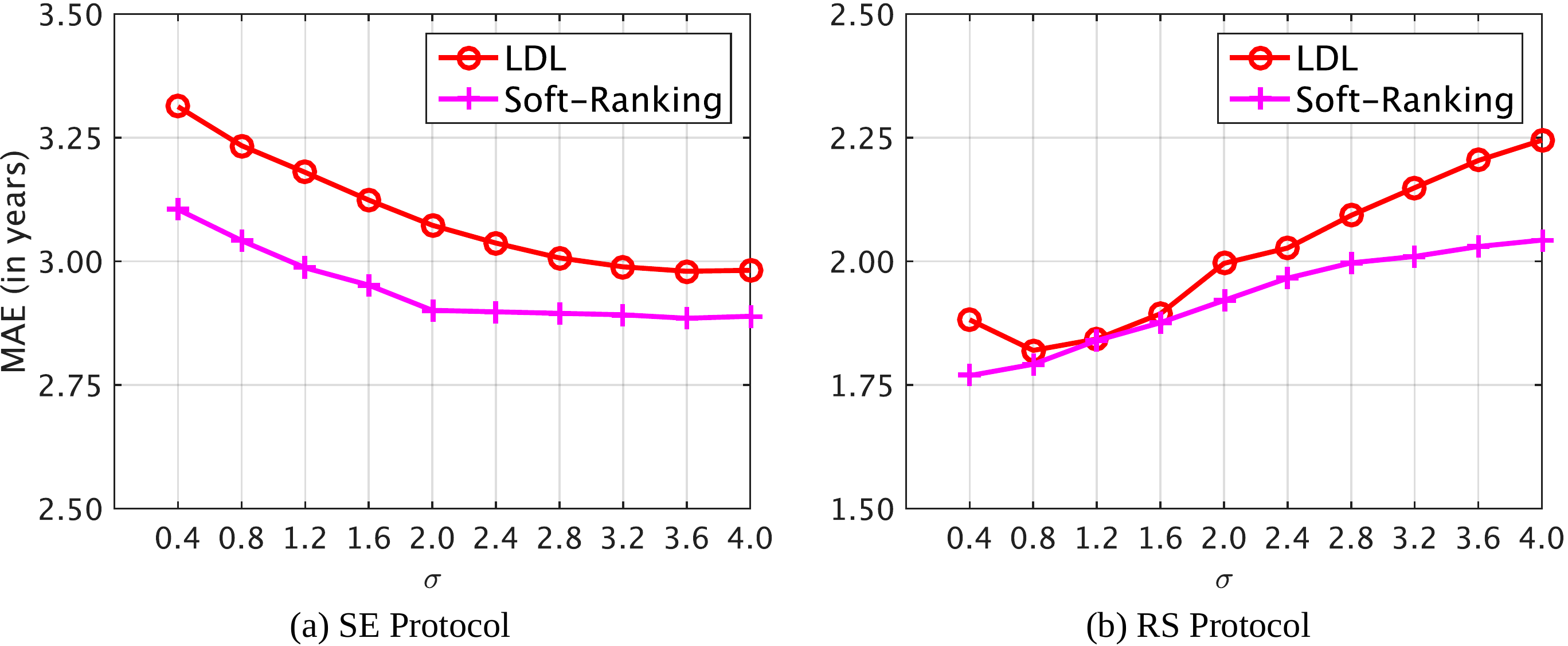}
\caption{Comparison in MAE between LDL and Soft-ranking with different $\sigma$ values.
(a) Performance on the SE protocol of Morph II;
(b) performance on the RS protocol of Morph II.
}
\label{tunesigma}
\end{figure}

We compare the performance of Soft-ranking with two popular age encoding strategies, $\emph{i.e.}$, LDL and Hard-ranking, which are implemented according to the methods outlined in Sec. \ref{Method}.
To facilitate a clean comparison, only the main task in Fig. \ref{pipeline} is utilized. Experiments are conducted on the SE protocol of Morph II and AgeDB, respectively.
There is no identity overlap between the training and testing sets for both experiments.
We carefully tune the $\sigma$ parameter of both LDL and Soft-ranking from 0.4 to 4.0 and report their best result, respectively.
The performance of LDL and Soft-ranking with different $\sigma$ values is illustrated in Fig.~\ref{tunesigma}.
The best performance of each age encoding method is summarized in Table \ref{comparison}.

From the results of the two experiments, it is clear that Soft-ranking consistently outperforms both the LDL and Hard-ranking methods.
This is because Soft-ranking combines two important properties of the facial age, while both LDL and Hard-ranking utilize only one of these facial age properties, respectively.
The above comparisons make a strong case for the effectiveness of the proposed age encoding method.

\begin{table}[!htbp]
\begin{center}
\begin{tabular}{|p{3.2cm}<{\centering}|p{1.2cm}<{\centering}|p{1.3cm}<{\centering}|p{1.2cm}<{\centering}|}
\hline
\multirow{2}{*}{Method} & Morph II (SE) & \multirow{2}{*}{AgeDB} & Morph II (RS) \\
\hline\hline
LDL          & 2.980           & 5.881 & 1.820          \\
Hard-ranking & 3.087           & 5.952 & \textbf{1.725} \\
Soft-ranking & \textbf{2.885}  & \textbf{5.741} & 1.769          \\
\hline
\end{tabular}
\end{center}
\caption{
Comparisons in MAE between different age encoding strategies.}
\label{comparison}
\end{table}

\subsubsection{Comparisons between different age encoding strategies on the RS protocol of Morph II}
\label{why}
We repeat the above experiments on the RS protocol of the Morph II database. Experimental results are also reported in Fig.~\ref{tunesigma} and Table \ref{comparison}.
It is interesting to note that Hard-ranking outperforms both Soft-ranking and LDL in this experiment.
This phenomenon can be explained as follows:
first, there is identity overlap between the training and testing sets in the RS protocol;
second, images of the same subject in the Morph II database are close in age value, meaning that the training and testing sets may be highly correlated.
As a result, a model that overfits on the training set may also perform well on the testing set.

Unlike LDL and Soft-ranking, which adopt soft labels to make use of the correlation between adjacent ages and thereby improve the generalization ability of deep models,
Hard-ranking forces classifiers to make predictions with high confidence.
Therefore, Hard-ranking creates a higher overfitting risk compared with the other two age encoding strategies.
The above phenomenon is also reflected in the optimal value of $\sigma$ for LDL and Soft-ranking:
a larger $\sigma$ encourages the model to emphasize the correlation between adjacent ages, while a smaller $\sigma$ enables the model to be more discriminative on the training set, but usually hurts its generalization ability.
In Fig.~\ref{tunesigma}, we show the performance of LDL and Soft-ranking with different $\sigma$ values under both RS and SE protocols of the Morph II database.
From this it can be seen that the optimal value of $\sigma$ for the RS protocol is significantly smaller than that for the SE protocol.

The above experiments indicate that the overlap in identity of the RS protocol may result in misleading results.
Therefore, we can conclude that the subject-exclusive (SE) protocol is more suitable for the evaluation of different age estimation algorithms.

\subsubsection{Effectiveness of Maskout regularization}
In this experiment, we investigate the effectiveness of Maskout using the two protocols of the Morph II database.
We also evaluate Maskout’s robustness against variations in the size of the erased area.
Soft-ranking is adopted as the age encoding strategy.
Experimental results are summarized in Table \ref{regularization}.
The baseline in this table refers to the model with only the main task in Fig. \ref{pipeline} is utilized.
These results demonstrate that Maskout can steadily improve age estimation accuracy,
which proves its effectiveness in regularizing the network to learn more robust parameters.

\begin{table}[tbp]
\begin{center}
\begin{tabular}{|p{3.2cm}<{\centering}|p{1.2cm}<{\centering}|p{1.3cm}<{\centering}|p{1.2cm}<{\centering}|}
\hline
\multirow{2}{*}{Method} & Morph II (RS) & Morph II (SE)\\
\hline\hline
baseline & 1.769&2.885 \\
Mask $3\times3$ & 1.696 & 2.837 \\
Mask $4\times4$ & \textbf{1.689} & \textbf{2.834} \\
Mask $5\times5$ & 1.698 &2.846 \\
\hline
\end{tabular}
\end{center}
\caption{
Comparisons in MAE with different settings of the Maskout regularization.}
\label{regularization}
\end{table}

\subsection{Comparisons with State-of-the-Art Methods}
To ensure fair comparison with existing works, we report the performance of our approach using one-step and two-step pre-training strategies, respectively.

\subsubsection{Real Age Estimation}
Real age estimation experiments are conducted on Morph II and AgeDB.
With reference to the ablation study experiments, we empirically set the $\sigma$ parameter in Eq. \ref{cdf1} as 0.4 for the experiment on the RS protocol of Morph II and as 3.6 for experiments on the other two protocols.

\textbf{Results on Morph II with RS protocol. }
Comparisons under this protocol are summarized in Table \ref{MorphII_RS}.
Most competitive approaches in Table \ref{MorphII_RS} adopt age encoding strategies \cite{author12, author13, author04, author28, author44, author72},
justifying the effectiveness of the existing age encoding methods. Our method achieves the best MAE performance of 1.67 years, which is lower than the current state-of-the-art MAE by a margin of 0.24 years.
Moreover, pre-training on the IMDB-WIKI database results in only minor performance improvements in this protocol, which indicates that our method can perform well even without perfect initialization.

\textbf{Results on Morph II with SE protocol. }
Compared with the RS protocol, the SE protocol of Morph II allows for no identity overlap between the training and testing sets;
therefore, age estimation is more challenging on this protocol.
As only a few works have adopted this protocol, there are limited results available to be used for comparison.
To facilitate fair comparison with existing works \cite{author31}, we report the results of our approach based on different network backbones,
$\emph{i.e.}$, VGG16 \cite{author92} and ResNet-34 \cite{author48}.
Comparisons in Table~\ref{MorphII_SE} show that our method outperforms the previous best result by a margin of 0.08 years.

\textbf{Results on AgeDB. }
The experimental results on Morph II are encouraging. However, all images in Morph II were captured in a controlled environment.
In the following, we further conduct experiments on AgeDB, which is one of the largest in-the-wild databases for age estimation.

Compared with other real age estimation databases, age estimation on AgeDB is significantly more challenging.
For example, Moschoglou $\emph{et al.}$~\cite{author91} tested the popular DEX model \cite{author27} on the entire AgeDB database and reported an MAE of 13.1 years.
This is partly because of the wide age span of AgeDB.
Since this paper is the first to report results on AgeDB using a standard cross-validation protocol, we are not able to compare our results with those of other works;
instead, we present the average performance of the proposed approach on the entire testing sets and their subsets of different age ranges in Table \ref{AgeDB}.

\subsubsection{Apparent Age Estimation}
We then compare the performance of the proposed method with recent approaches on the apparent age estimation task.
We also compare the performance of LDL, Hard-ranking, and Soft-ranking in Tables~\ref{CLAP15} and~\ref{CLAP16}.
To facilitate fair comparison, experimental settings for the three age encoding methods are exactly the same.
Experimental results with the three age encoding methods are denoted as `Ours (LDL)', `Ours (Hard-ranking)', and `Ours (Soft-ranking)', respectively.
As the value of $\sigma$ for each image is available for both the CLAP2015 and CLAP2016 databases,
we simply set $\sigma$ as the value provided by the databases for both the main task and the auxiliary tasks.

\textbf{Results on CLAP2015.}
Performance comparison on this database is summarized in Table \ref{CLAP15}.
From these results, it can be seen that Soft-ranking outperforms the other two age encoding methods.
Besides, the proposed method achieves the best $ \epsilon$-error performance of 0.232.
In particular, our method significantly outperforms AGEn \cite{author44},
which is an ensemble system including eight networks and employs complex data augmentation during the testing stage.
In comparison, we employ only a single model and light data augmentation in testing.
Moreover, our approach outperforms \cite{author76} by as much as 0.045 in terms of $\epsilon$-error while using same backbone model, $\emph{i.e.}$, ResNet-34.
In addition, the proposed approach still yields a clear performance advantage even when \cite{author76} adopts a more powerful backbone architecture, $\emph{i.e.}$, RoR-34.
In summary, the above comparisons demonstrate the effectiveness of the proposed methods.

\textbf{Results on CLAP2016. }
Comparisons on this database are tabulated in Table \ref{CLAP16}.
It is shown that Soft-ranking outperforms the other two age encoding methods, which is consistent with the experimental results on CLAP2015.
Among the existing approaches, the method proposed in \cite{author28} outperforms the others by a large margin.
This is partly because it employs both private training data and a manually cleaned IMDB-WIKI database;
moreover, it adopts a multi-model ensemble to further improve the performance.
In comparison, we employ only publicly available data for training and a single model for prediction.
Nevertheless, our proposed approach still outperforms \cite{author28}.
To the best of our knowledge, this is the first approach to achieve better performance than \cite{author28} after the CLAP2016 challenge.

\begin{table}
\begin{center}
\begin{tabular}{|p{3.5cm}<{\raggedright}|p{1.2cm}<{\centering}|p{2cm}<{\centering}|p{2cm}<{\centering}|}
\hline
Method & MAE \\
\hline\hline
OHRank \cite{author42} & 6.07 \\
CNN+ELM \cite{author75} & 3.44 \\
OR-CNN \cite{author12} & 3.27 \\
SSR-Net \cite{author52} & 3.16 \\
DEX \cite{author27} & 3.25 \\
DEX$^{*}$ \cite{author27} & 2.68 \\
LSDML \cite{author20} & 3.08 \\
M-LSDML \cite{author20} & 2.89 \\
SAF \cite{author93} & 2.97 \\
Ranking-CNN \cite{author13} & 2.96 \\
CMT w/ GM, LE \cite{author53} & 2.89 \\
C3AE \cite{author89} & 2.78 \\
C3AE$^*$ \cite{author89} & 2.75 \\
RGAN$*$ \cite{author74} & 2.61\\
AGEn \cite{author44} & 2.93 \\
AGEn$^*$ \cite{author44} & 2.52 \\
AL-RoR-34$^*$ \cite{author76} & 2.36\\
VGG-16 CNN + LDAE$^*$ \cite{author28} & 2.35 \\
mean-variance loss \cite{author31} & 2.41 \\
mean-variance loss$^*$ \cite{author31} & 2.16 \\
DLDL-v2 \cite{author04} & 1.97 \\
CEN$^*$ \cite{author72} & 1.91 \\\hline\hline
Ours & 1.69 \\
Ours$^*$ & \textbf{1.67} \\
\hline
\end{tabular}
\end{center}
\caption{Comparisons in MAE between our approach and state-of-the-art methods on the Morph II database (RS protocol).
$*$ indicates that the model has been pre-trained on the IMDB-WIKI dataset. }
\label{MorphII_RS}
\end{table}

\begin{table}
\begin{center}
\begin{tabular}{|p{3.5cm}<{\raggedright}|p{1.2cm}<{\centering}|p{2cm}<{\centering}|p{2cm}<{\centering}|}
\hline
Method & MAE \\
\hline\hline
mean-variance loss~\cite{author31} & 2.80 \\
mean-variance loss$^*$~\cite{author31} & 2.79 \\\hline\hline
Ours (ResNet-34) & 2.83 \\
Ours (ResNet-34)$^*$ & 2.83 \\
Ours (VGG16)$^*$ & \textbf{2.71} \\
\hline
\end{tabular}
\end{center}
\caption{Comparisons in MAE between our approach and state-of-the-art methods on the Morph II database (SE protocol).
~\cite{author31} adopts the VGG16 backbone model.
}
\label{MorphII_SE}
\end{table}

\begin{table}[tp]
\begin{center}
\begin{tabular}{|p{3.5cm}<{\raggedright}|p{1.2cm}<{\centering}|p{2cm}<{\centering}|p{2cm}<{\centering}|}
\hline
Method (Age Range) & MAE \\
\hline\hline
Ours$^*$ (1-19)   & 6.519 \\
Ours$^*$ (20-39)  & 4.849 \\
Ours$^*$ (40-59)  & 5.744 \\
Ours$^*$ (60-79)  & 5.905 \\
Ours$^*$ (80-101) & 8.816 \\\hline\hline
Ours$^*$ (1-101)  & 5.581 \\

\hline
\end{tabular}
\end{center}
\caption{Result of our approach on the AgeDB database. }
\label{AgeDB}
\end{table}

\begin{table}[!htbp]
\begin{center}
\begin{tabular}{|p{3.5cm}<{\raggedright}|p{1.2cm}<{\centering}|p{2cm}<{\centering}|p{2cm}<{\centering}|}
\hline
Method & $\epsilon$-error & Single Model? \\
\hline\hline
DLDL \cite{author37} & 0.310 & YES \\
DEX$^*$ \cite{author27} & 0.282 & YES \\
DEX$^*$ \cite{author27} & 0.265 & NO \\
DLDL-v2 \cite{author04} & 0.277 & YES \\
AGEn$^*$ \cite{author44} & 0.264 & NO \\
AL-ResNets-34$^*$ \cite{author76} & 0.277 & YES \\
AL-RoR-34$^*$ \cite{author76} & 0.255 & YES \\
BridgeNet$^*$ \cite{author90} & 0.255 & YES \\\hline\hline
Ours (LDL)& 0.250 & YES \\
Ours (Hard-ranking)& 0.254 & YES \\
Ours (Soft-ranking)& 0.244 & YES \\
Ours (Soft-ranking)$^*$ & \textbf{0.232} & YES \\
\hline
\end{tabular}
\end{center}
\caption{Comparisons in $\epsilon$-error between our approach and state-of-the-art methods on the CLAP2015 database. }
\label{CLAP15}
\end{table}

\begin{table}[!htbp]
\begin{center}
\begin{tabular}{|p{3.5cm}<{\raggedright}|p{1.2cm}<{\centering}|p{2cm}<{\centering}|p{2cm}<{\centering}|}
\hline
Method & $\epsilon$-error & Single Model? \\
\hline\hline
SAF \cite{author93} & 0.374 & NO \\
AGEn$^*$ \cite{author44} & 0.310 & NO \\
DADL \cite{author45} & 0.321 & NO \\
mean-variance loss$^*$ \cite{author31} & 0.287 & YES \\
AL-RoR-34$^*$ \cite{author76} & 0.286 & YES \\
DLDL-v2 \cite{author04} & 0.267 & YES \\
VGG-16 CNN + LDAE$^*$ \cite{author28} & 0.241 & NO \\\hline\hline
Ours (LDL)& 0.251 & YES \\
Ours (Hard-ranking)& 0.254 & YES \\
Ours (Soft-ranking)& 0.246 & YES \\
Ours (Soft-ranking)$^*$ & \textbf{0.232} & YES \\

\hline
\end{tabular}
\end{center}
\caption{Comparisons in $\epsilon$-error between our approach and state-of-the-art methods on the CLAP2016 database. }
\label{CLAP16}
\end{table}

\section{Conclusion}
\label{sectionConclusion}
In this work, we aim at overcoming the challenges associated with facial age estimation by making multiple contributions.
First, a new age encoding method is proposed, which we have named Soft-ranking. Soft-ranking seamlessly combines the ordinal information and the correlation among adjacent ages,
and therefore, has notable advantages compared with existing age encoding strategies.
Second, we propose the novel Maskout strategy, which regularizes deep networks in order to learn more robust facial representations for age estimation.
Third, we point out that the identity overlap between the training and testing sets may cause misleading results during the evaluation of age estimation algorithms.
Based on the proposed methods, we obtain state-of-the-art results on three popular age estimation databases.
In particular, our approach is the first single network model to break the record set by Antipov $\emph{et al.}$ \cite{author28} in the CLAP2016 competition.

{\small
\bibliographystyle{IEEEtran}
\bibliography{reference_using_Abbreviations}
}

\end{document}